\pdfoutput=1
\documentclass[sn-mathphys]{sn-jnl}% Math and Physical Sciences Reference Style
\jyear{2021}%

\theoremstyle{thmstyleone}%
\theoremstyle{thmstyletwo}%

\theoremstyle{thmstylethree}%
\usepackage[numbers,sort&compress]{natbib}
\usepackage{hyperref}
\usepackage{graphicx}
\usepackage{booktabs}
\usepackage{array}
\usepackage{multirow}
\usepackage{longtable}
\usepackage{rotating}
\usepackage{overpic}
\usepackage{textcomp}
\usepackage{wrapfig}
\usepackage{amsthm,amsmath,amssymb}
\usepackage{mathrsfs}
\makeatletter
\newcommand*\bigcdot{\mathpalette\bigcdot@{.5}}
\newcommand*\bigcdot@[2]{\mathbin{\vcenter{\hbox{\scalebox{#2}{$\m@th#1\bullet$}}}}}
\makeatother

\begin{document}

\title[CI-Net: a Joint Task learning Network using Contextual Information]{CI-Net: Contextual Information for Joint Semantic Segmentation and Depth Estimation}


\author[1]{\fnm{Tianxiao} \sur{Gao}}\email{201920116403@scut.edu.cn}

\author*[1]{\fnm{Wu} \sur{Wei}}\email{eeweiwu@126.com}
%%\equalcont{These authors contributed equally to this work.}

\author[1]{\fnm{Zhongbin} \sur{Cai}}\email{201921017250@scut.edu.cn}

\author*[2]{\fnm{Zhun} \sur{Fan}}\email{zfan@stu.edu.cn}

\author[3]{\fnm{Shengquan} \sur{Xie}}\email{s.q.xie@leeds.ac.uk}

\author[4]{\fnm{Xinmei} \sur{Wang}}\email{wangxm@cug.edu.cn}

\author[1]{\fnm{Qiuda} \sur{Yu}}\email{yuqiuda@163.com}
%%\equalcont{These authors contributed equally to this work.}

\affil*[1]{\orgdiv{School of Automation Science and Engineering}, \orgname{South China University of Technology}, \orgaddress{\city{Guangzhou}, \postcode{510000}, \state{Guangdong}, \country{China}}}

\affil[2]{\orgdiv{College of Engineering}, \orgname{Shantou University}, \orgaddress{\city{Shantou}, \postcode{515000}, \state{Guangdong}, \country{China}}}

\affil[3]{\orgdiv{School of Electronic and Electrical Engineering}, \orgname{University of Leeds}, \orgaddress{\city{Leeds}, \postcode{LS2 9JT}, \state{UK}, \country{China}}}

\affil[4]{\orgdiv{School of Automation}, \orgname{China University of Geosciences}, \orgaddress{\city{Wuhan}, \postcode{430074}, \state{Hubei}, \country{China}}}


\abstract{Monocular depth estimation and semantic segmentation are two fundamental goals of scene understanding. Due to the advantages of task interaction, many works study the joint task learning algorithm. However, most existing methods fail to fully leverage the semantic labels, ignoring the provided context structures and only using them to supervise the prediction of segmentation split, which limit the performance of both tasks. In this paper, we propose a network injected with contextual information (CI-Net) to solve the problem. Specifically, we introduce self-attention block in the encoder to generate attention map. With supervision from the ideal attention map created by semantic label, the network is embedded with contextual information so that it could understand scene better and utilize correlated features to make accurate prediction. Besides, a feature sharing module is constructed to make the task-specific features deeply fused and a consistency loss is devised to make the features mutually guided. We evaluate the proposed CI-Net on the NYU-Depth-v2 and SUN-RGBD datasets. The experimental results validate that our proposed CI-Net could effectively improve the accuracy of semantic segmentation and depth estimation.}




\maketitle

\section{Introduction}\label{sec1}

Scene understanding is an important yet challenging task in computer vision, which has significant contribution to visual simultaneous localization and mapping(vSLAM) system \cite{Kang_2010}, robot navigation \cite{Husbands_2021}, autonomous driving \cite{Lee_2020} and other applications. Two fundamental goals of scene understanding are monocular depth estimation \cite{eigen2014depth,Xu_2018,Liu_2015,Cao_2018} and semantic segmentation \cite{Wang_2018, Lan_2021,Long_2015,Lin_2016}, which have been extensively researched by utilizing deep learning. Recently, some works \cite{Xu_2018,guizilini2020semantically,Zhang_2019} noticed the interactions between these two tasks and utilized the common characteristics to improve each other, which achieved great performance. However, most of them used a deep structure as encoder such as ResNet-101 \cite{guizilini2020semantically}, ResNet-50 \cite{Zhang_2018}, SE-ResNet \cite{choi2020safenet}, introducing a large number of downsampling and stride operations, which had negative influence \cite{Fu_2018} towards depth estimation and semantic segmentation where fine-grained information is crucial. Despite there are also some works adopting strategies like skip-connection \cite{choi2020safenet}, up-projection \cite{Zhang_2019} and multi-scale training loss \cite{Zhang_2018} to mitigate this problem, these schemes have great demands on computation and memory.

Another shortcoming is that current works about joint learning didn't fully exploit the contextual information of the semantic labels. As far as we know, most of them simply utilized the labels to supervise the predictions of semantic and depth splits, making limited contribution to scene understanding of the network. In \cite{Chen_2021}, Chen et al. pointed out that how to obtain the correlation of inter-object and intra-object is crucial for depth estimation. Yu et al. \cite{Yu_2020} also argued such context make feature representation more robust for semantic segmentation. Therefore, could we excavate the information of labels more deeply to assist the network modeling such correlation? Moreover, most approaches achieve task interaction through adding pixel-wise features \cite{Jiao_2018,choi2020safenet}, simply sharing encoder commonly \cite{Klingner_2020} or sharing parameters of convolutional layers \cite{Zhang_2019}. Although these methods leveraged the correlation between tasks, their ways of fusing features were rough. For example, in \cite{choi2020safenet}, they fused the features via direct weighted addition and then added them to task-specific features. This simple structure may make the network difficult to learn more useful representations of the shared features.

To overcome the mentioned problems, this paper presents a network injected with contextual information (CI-Net). We adopt the dilated residual structure where the dilation operation replaces a part of downsampling layers, guaranteeing large receptive fields and avoiding introducing unnecessary parameters. To fully leverage the provided context of semantic labels, we plug a scene understanding module (SUM) with contextual supervision, which captures the similarity of pixels belong to the same classes and difference of those pertaining to different classes. Specifically, we introduce self-attention block \cite{Wang_2018,Shaw_2018} to generate attention map and exploit the semantic labels to create the ideal attention map indicating whether pair of pixels belongs to the same classes or not. The attention training loss injects the contextual prior into the network, which makes the structure know to use correlated features for more accurate prediction. To make these two tasks deeply interacted, we present two approaches. The first one is that we design a feature sharing module (FSM). Instead of simply adding the task-specific features, we concatenate and put them through a series of downsampling and upsampling operations for more useful representations being obtained. We also devise a consistency loss between the depth and semantic features, forcing them to maintain the intrinsic consistency of first-order relationship.

To summarize, the contributions of this paper are in three aspects:

\begin{itemize}
	\item [--]
	We propose a dilated network embedded a scene understanding module with contextual supervision to inject contextual prior about the correlation of inter-class and intra-class, predicting both the depth and semantic segmentation maps.
	\item [--]
	We construct a feature sharing module to deeply fuse the task-specific features and put forward a consistency loss to keep the respective features consistent in the relationship with adjacent features. 
	\item [--]
	Extensive experiments are performed to demonstrate the effectiveness of our methods. And the proposed model achieves competitive results against other approaches of depth estimation and semantic segmentation on NYU-Depth-v2 and SUN-RGBD datasets. 
\end{itemize}

\section{Related Works}\label{sec2}

\subsection{Monocular Depth Estimation}
The first research about predicting monocular depth is Saxena's work \cite{saxena2005learning}, which introduced Markov Random Field (MRF) to exploit the geometic priors. Later on, with the appearance of convolutional neural network (CNN), Eigen et al. \cite{eigen2014depth} proposed a model intergrating the global and local information of a single image. Since then, a large number of studies using deep learning methods have emerged. Liu et al. \cite{Liu_2015} combined Convolutional Neural Fields (CRF) with fully convolutional networks and put foward a superpixel pooling method making the inference faster. Laina et al. \cite{Laina_2016} proposed a fully convolutional residual architecture and four up-sampling models to restore the resolution of depth map. To reduce the information loss induced by excessive pooling, Fu et al. \cite{Fu_2018} employed the atrous spatial pyramid pooling (ASPP) and presented an ordinal loss to model the depth prediction as an ordinal regression problem. Inspired by \cite{Fu_2018}, Chen et al. \cite{Chen_2021} proposed soft ordinal inference to exploit the predicted probabilities of the whole depth intervals and replaced ASPP with self-attention module to capture the global context. Recently, Yin et al. \cite{Yin_2019} projected the depth map to obtain the 3D point cloud, exploiting the loss between virtual normal and ground truth to train the model, which significantly improved the accuracy. There were also some works  \cite{Yin_2018,Yang_2018,Godard_2017} who employed the geometric constraints of the consective image sequence to complete unsupervised depth prediction. 

\subsection{RGB-D Semantic Segmentation}
The outstanding work proposed by Long et al. \cite{Long_2015}, Fully Convolutional Network (FCN) achieved great improvement of semantic segmentation. Since then, many scholars \cite{Yuan_2020,Yuan_2021,Yu_2020,Lin_2016,kendall2015bayesian} researched on this scene understanding task using single RGB images. After the RGB-D dataset was released, some approaches \cite{Qi_2017,Hazirbas_2017,He_2017} discovered that fusing depth images could significantly improve the segmentation results. Recently, Hu et al.\cite{Hu_2019_AcNet} proposed Attention Complementary Network which fused weighted depth and semantic features in the encoder. The fusion implementation enabled ACNet to exploit more high-quality features. Hung et al.\cite{Hung_2019} designed LDFNet, which is also a fusion-based network. Its novelty of incorporating luminance, depth and color information made great success in semantic segmentation. To reduce the inference time, Chen et al.\cite{Chen_2021_Spatial} proposed Spatial information guided Convolution (S-Conv) which extracted geometric information as convolutional weights and infers the sampling offset according to the 3D spatial information. Different from these algorithms who aimed to improve semantic segmentation with the facilitation of RGB-D images, we design a joint-task learning network to boost both depth estimation and semantic segmentation with only RGB images as input through the deep interaction of each task.

\subsection{Joint Semantic Segmentation and Depth Estimation}
Due to the common character of pixel-level prediction among different tasks, some works paid attention to studying joint learning. In \cite{Peng_Wang_2015}, Wang et al. used a network to jointly predict a depth map and a semantic probability map. The network, however, is limited by only the last layer being changed for prediction. Jiao et al. \cite{Jiao_2018} proposed a network with a backbone encoder and two sub-networks as decoders for respective prediction, increasing both the accuracy of depth estimation and semantic segmentation dramatically. Later, PAD-Net was proposed by Xu et al. \cite{Xu_2018_Pad}, which used four intermediate auxiliary tasks, providing abundant information for prediction. In the recent research, the SOSD-Net \cite{He_2021} made full use of the geometric relations between depth estimation and semantic segmentation to predict. Although these works achieved outstanding performance, they didn't exploit the feature that semantic labels could help network capture prior contextual knowledge of the scene to improve the accuracy of prediction.

\subsection{Attention Mechanism}
Xu et al. \cite{xu2015show} was the pioneer who innovated attention mechanism into deep learning. Since then, a profusion of attention methods were designed mainly including channel attention \cite{Hu_2018}, spatial attention \cite{jaderberg2015spatial} and self-attention \cite{Wang_2018}. Inspired by them, some works that incorporated attention mechanism into depth prediction have emerged. Chen et al. \cite{Chen_2020} interpolated channel attention block into the encoder and spatial block into the decoder to avoid losing structural information. Xu et al. \cite{Xu_2018} presented a fused CRF model guided by multi-scale attention. Choi et al. \cite{choi2020safenet} designed an affinity propagation unit to make the depth features guided by semantic attention map.

In this work, we present a model jointly learning semantic and depth representations. We introduce a shared attention block for these two tasks with contextual supervision so that the network can understand the scene better for prediction. Moreover, we design feature sharing modules to combine the semantic and depth features, making use of the task-wise information. Notice that the similarity and discrepancy of the two kinds of features should keep consistent , we also construct a novel consistency loss.

\section{Methods}\label{sec3}

This section illustrates our proposed method for joint semantic segmentation and depth estimation from single RGB image. The first three subsections introduce the architecture of our proposed CI-Net and its sub-modules. The last subsection outlines the training losses. 

\subsection{Network Architecture}
\begin{figure*}[htbp]
	\centering
	\includegraphics[width=4.7in]{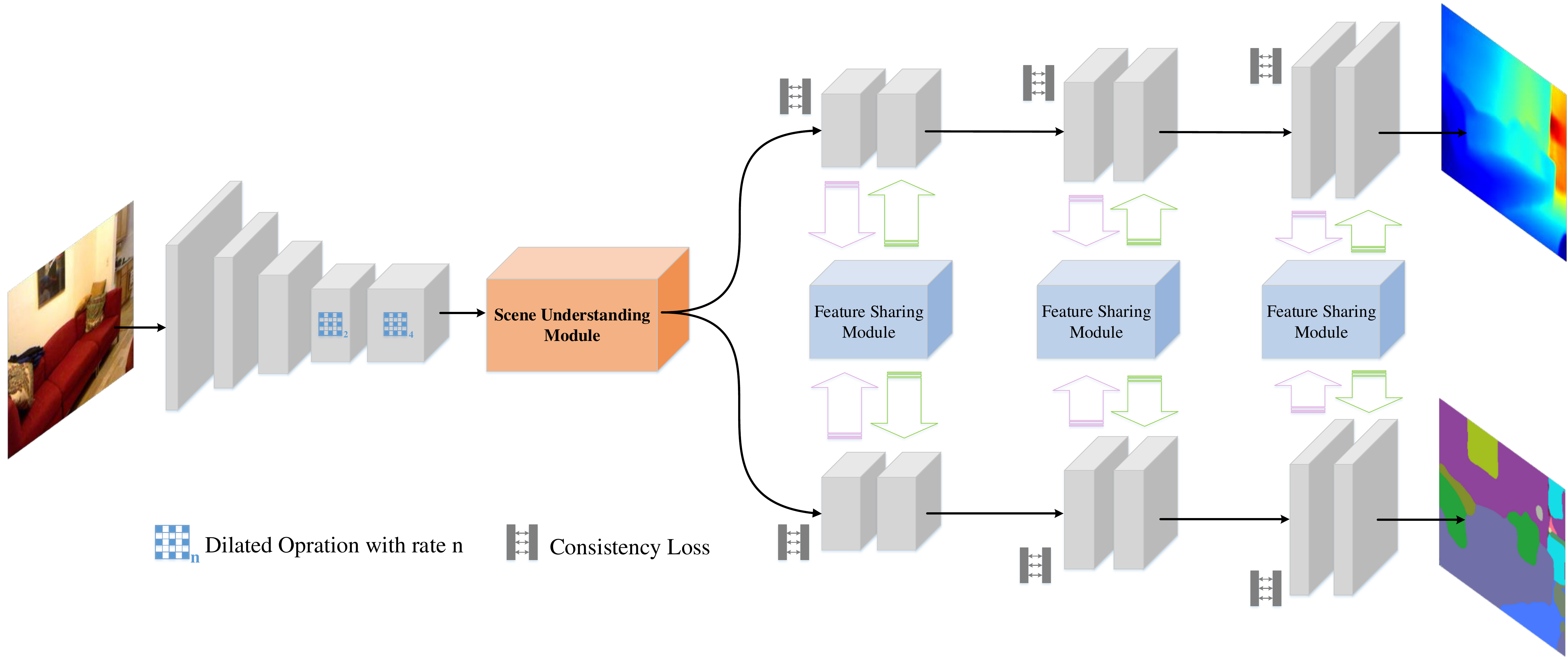}
	\caption{The overview of our CI-Net for joint depth estimation and semantic segmentation. We adopt dilated operation in the backbone to mitigate the harm of over-downsampling. At the end of the encoder, the SUM is designed to aggregate the contextual information. Then the network breaks into depth and segmentation split. To deeply fuse the task-specific features, the FSM is proposed. Finally, a consistency loss is formulated to make the depth and segmentation features mutually guided.}
	\label{network architecture}
\end{figure*}

The proposed Contextual Information Network (CI-Net) uses the encoder-decoder scheme as shown in Figure \ref{network architecture}. For the encoder, we choose the ResNet \cite{He_2016} for its identity mapping tackling the vanishing gradient problem in deeper network. Another benefit of ResNet is its large receptive field \cite{Laina_2016}, which is a crucial factor for depth estimation and semantic segmentation. However, instead of deploying ResNet as encoder directly like \cite{guizilini2020semantically,Zhang_2018,choi2020safenet}, we also adopt dilation strategy \cite{Yu_2017} to mitigate the negative effect of overdownsampling in ResNet which may hinder the predictions of fine-grained depth and semantic maps. With the last two 2-dilated and 4-dilated residual blocks, the original resolution is lowered to $1/8$ instead of $1/32$, reducing the detailed information loss. 

In the decoder, a scene understanding module (SUM) with supervised attention block is designed to fully exploit the semantic labels to obtain context prior of inter-class and intra-class, which benefits the model to understand scene better for later prediction. The network then breaks apart into two branches for estimating depth and segmenting semantics. During this stage, we present a feature sharing module (FSM) to share feature representations so that these two splits could fully exploit different levels of features. And a consistency loss is formulated to keep the task-specific features consistent. More details about these methods are described in the following subsections.
%\IEEEpubidadjcol

\subsection{Scene Understanding Module}
Our motivation mainly comes from two folds: i) Pixels of same objects tend to have continuous or similar depth values while the depths of different objects have large discrepancies. ii) Under the background joint task learning of semantic segmentation and depth estimation, semantic labels contain information of each class so that it's easy to know whether pixels belong to the same classes or not. Thus our goal is to find an effective way to make the network have prior knowledge of the categorical relationship. To achieve that, we utilize the semantic labels for supervising the network with an attention loss to capture the correlation of the pixels belonging to same classes and the distinction of different classes. With the prior knowledge of the scene, the profitable information for prediction could be searched in a limited, related space instead of the whole region. Then the depth split, on the one hand, will be prevented from capturing the unrelated features. For example, the region of sky should not be used to predict the depth of ground and this behavior is hindered by the context prior for the gap between these two objects. On the other hand, the semantic one also benefits since it makes better judgments from the information of inter-class and intra-class. We encapsulate this process of obtaining contextual information in the scene understanding module, which is illustrated in the following content.

\begin{figure}[htbp]
	\centering
	\includegraphics[width=3.5in]{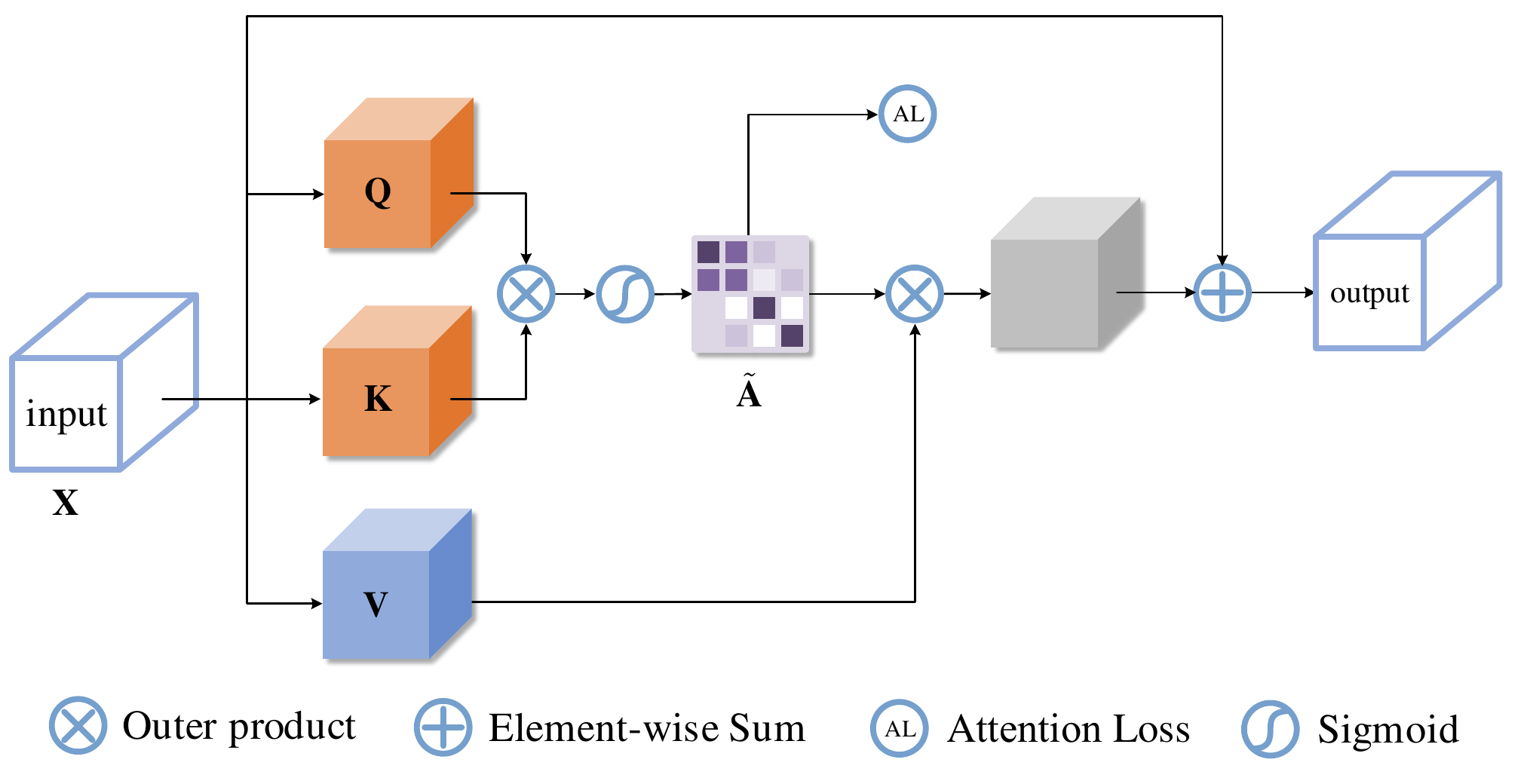}
	\caption{The structure of SUM. The generated attention map captures context prior of inter-class and intra-class so that the network understand the scene better.}
	\label{scene understanding module}
\end{figure}

The architecture of scene understanding module is presented in Figure \ref{scene understanding module}. Similar to non-local block \cite{Wang_2018}, it first uses $1 \times 1$ convolutions to transform the input features $\mathbf{X} \in R^{N \times C_{f}}$ into query,key and value results represented by $\mathbf{Q} \in R^{N \times C_{q}}$, $\mathbf{K} \in R^{N \times C_{k}}$ and $\mathbf{V} \in R^{N \times C_{v}}$ respectively, where $N=H \times W$ is the resolution. Then the predicted attention map $\tilde{\mathbf{A}}$ can be obtained with
\begin{equation}
	\tilde{\mathbf{A}}=Sigmoid(\mathbf{Q}^{T}\mathbf{K}),
\end{equation}
where $Sigmoid(\bigcdot)$ is the sigmoid function, which ensures the attention values are in range $[0,1]$. After that, the value results are multiplied with the attention map to capture the correlation with each pixel. By finally being added a skip connection to avoid the problem of vanishing gradient, the output $\mathbf{Y} \in R^{N \times C_{f}}$ can be defined as
\begin{equation}
	\mathbf{Y}=\tilde{\mathbf{A}}\mathbf{V}+\mathbf{X}.
\end{equation}

\begin{figure}[htbp]
	\centering
	\includegraphics[width=3.55in]{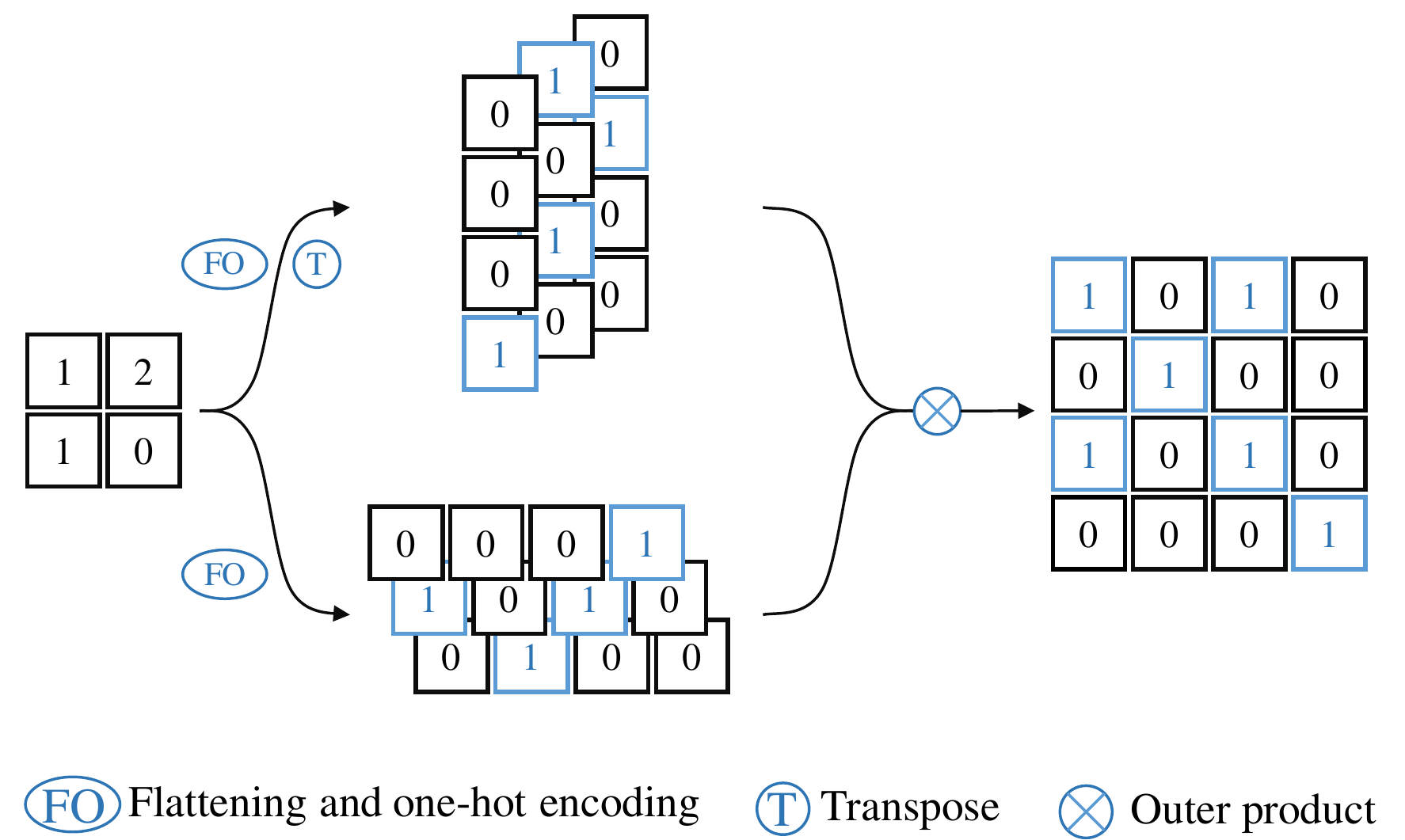}
	\caption{The process of generating the ideal attention map. First we implement one-hot encoding for the semantic label and then flatten it into a $HW \times 1 \times C$ matrix $\mathbf{M}$ where $C$ denotes the number of categories. After the operation of outer product, the ideal attention map $\mathbf{A}$ is constructed with size of $HW \times HW$. It can be noticed that the pixels of same classes in semantic label have the same value $1$ in the attention map.}
	\label{one-hot encoding}
\end{figure}

To capture the context prior of pixels, we adopt the method of \cite{Yu_2020} to generate the ideal attention map. As it can be seen in Figure \ref{one-hot encoding}, given the ground truth , we can know the label of each pixel. To transform it into the relation between different pixels, the ground truth is first down-sampled into the size of $H\times W$ and then flattened into a vector $\mathbf{m}$ of size $1\times N$. After the one-hot encoding implementation, new binary columns which indicate the presence of value from the ground truth are created, leading to a $H\times W\times C$ matrix $\mathbf{M}$, where $C$ represents the number of total categories. The matrix $\mathbf{M}$ is then reshaped into size of $N\times C$ and finally the ideal attention map $\mathbf{A}$ is constructed with
\begin{equation}
	\mathbf{A}=\mathbf{M}\mathbf{M}^{T}.
\end{equation}
It's clear that in the ideal attention map, pixels of same classes are labeled as 1, and 0 otherwise, which aggregates the contextual information of intra-class and inter-class. And we employ the binary cross entropy loss as the attention loss:
\begin{equation}
	\mathcal{L}_{att}=-\sum_{i,j}(\mathbf{A}_{i,j}log\tilde{\mathbf{A}}_{i,j}+((1-\mathbf{A}_{i,j})log(1-\tilde{\mathbf{A}}_{i,j})),
\end{equation}
where $A_{i,j}$ denotes the pixel at the location $(i,j)$ of the predicted attention map. 

It is worth noting that we utilize semantic labels instead of depth to inject context prior. One reason is that it is difficult to find a feasible and suitable representation of depth context. Although there exist some works proposing to use Kullback-Leibler divergence \cite{Chen_2021} or planar structures \cite{Huynh_2020}, their methods are limited to only depth estimation task. But for joint task learning the correlation of different objects is harmful to semantic segmentation.

\begin{figure}[htbp]
	\centering
	\begin{overpic}[width=3.1in]{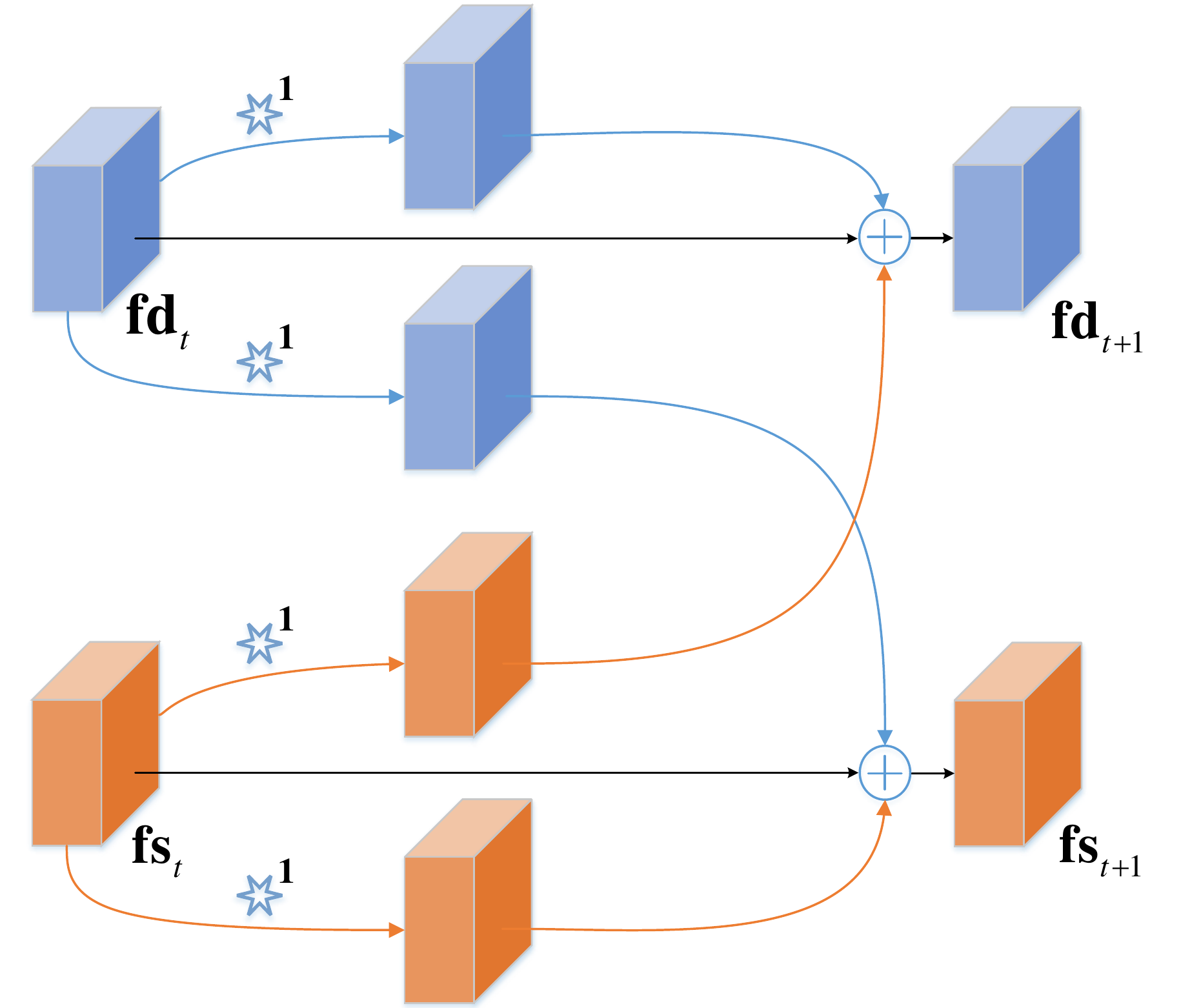}
		\put(1,82){(a)}
	\end{overpic}\\
	\begin{overpic}[width=3.5in]{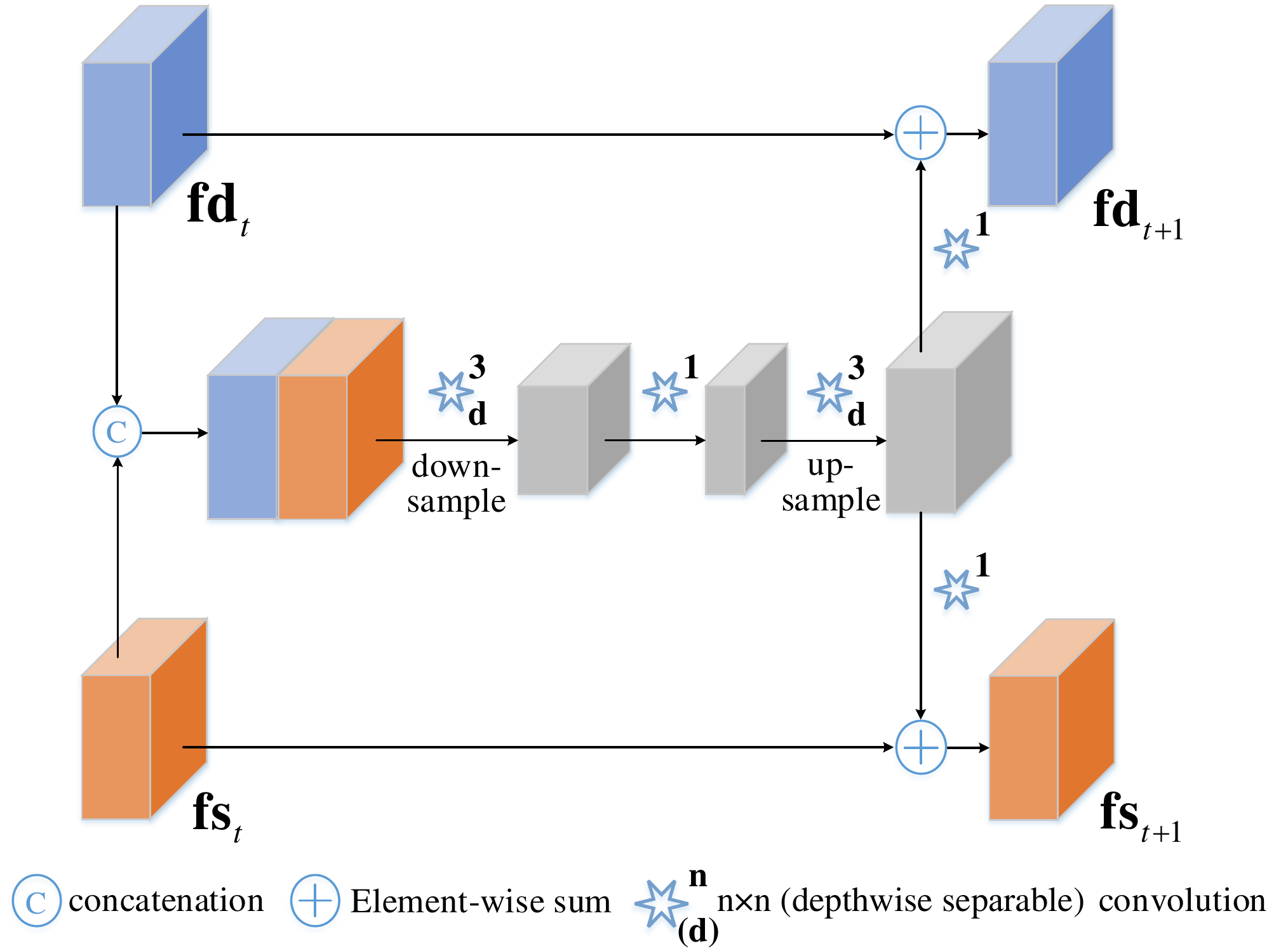}
		\put(6,75){(b)}
	\end{overpic}
	\caption{(a) The architecture of LSU; (b) The architecture of our proposed FSM \cite{choi2020safenet,Jiao_2018}. our module can capture extensive interacted context, while the comparative one only captures local interaction.}
	\label{feature sharing modules}
\end{figure}

\subsection{Feature sharing Module}
In the decoder, the network splits into depth and semantic branches. We design a feature sharing module (FSM) aiming to make two branches share the features with each other so that they can take full advantage of semantic and depth information. The structure of FSM is presented in Figure \ref{feature sharing modules}(b). The depth features $\mathbf{fd}_{t}$ and segmentation features $\mathbf{fs}_{t}$ are first concatenated, then fed into an architecture resembling encoder-decoder. We utilize $C(\bigcdot)$ to represent the series of convolutions in the aforementioned process. It can be noticed that we use depthwise separable convolution to reduce computational cost. Eventually the commonly shared features are allocated adaptively into two branches via $1\times 1$ convolutions $C_{\mathbf{fd}}^{1\times1}(\bigcdot)$, $C_{\mathbf{fs}}^{1\times1}(\bigcdot)$. Followed by residual connections, the features $\mathbf{d}_{t+1}$ and $\mathbf{s}_{t+1}$ can be obtained by:
\begin{equation}
	\begin{split}
		&\mathbf{fd}_{t+1} = \mathbf{fd}_{t} + C_{\mathbf{fd}}^{1\times1}(C(concat(\mathbf{fd}_{t},\mathbf{fs}_{t}))), \\
		&\mathbf{fs}_{t+1} = \mathbf{fs}_{t} + C_{\mathbf{fs}}^{1\times1}(C(concat(\mathbf{fd}_{t},\mathbf{fs}_{t}))).
	\end{split}
\end{equation}
We also compare our structure with lateral sharing unit (LSU) proposed in \cite{choi2020safenet,Jiao_2018}, which is shown in Figure \ref{feature sharing modules}(a). It can be seen that the task-shared features $\mathbf{fs}_{t+1}$ and $\mathbf{fd}_{t+1}$ are obtained with summation of three features, which can be formulated by:
\begin{equation}
	\begin{split}
		\mathbf{fd}_{t+1} = \mathbf{fd}_{t} + C_{\mathbf{fd}1}^{1\times1}(\mathbf{fd}_{t}) +  C_{\mathbf{fs}2}^{1\times1}(\mathbf{fs}_{t}), \\
		\mathbf{fs}_{t+1} = \mathbf{fs}_{t} + C_{\mathbf{fd}2}^{1\times1}(\mathbf{fd}_{t}) +  C_{\mathbf{fs}1}^{1\times1}(\mathbf{fs}_{t}).
	\end{split}
\end{equation}
Although their method, to some extend, realizes the interaction of different features, providing information for later predictions, we argue that the element-wise summation can only obtain local information, making limited use of the fused features. For example, the depth features located at $(i,j)$ can only sum with the corresponding semantic features. Contrast to LSU, our method implements sampling towards the interacted features, which encapsulates large area of features and augments the presentation ability. Therefore, each task-specific feature acquires more useful information. We insert FSM before each stage of upsampling, benefiting depth split and segmentation split to exploit multi-level fused information.

\subsection{Training Loss Function}
Besides the previously mentioned attention loss, our loss function includes three other parts: consistency loss, depth loss and segmentation loss.

\textbf{Consistency Loss:} Inspired by \cite{guizilini2020semantically}, we design a consistency loss to make semantic and depth branches guide each other mutually. Specifically, the features that are distinct or similar in semantic feature map should keep the same nature as in depth representations. For example, the semantic features of the sky and tree are extremely different since they belong to different classes, while the corresponding two depth features are also discrepant because the distances of them have large gap. Therefore, we employ this character to supervise the consistency loss of task-specific features, the form of which is defined as
\begin{equation}
	\begin{split}
		&\mathcal{L}_{con}=\sum_{l}\sum_{i,j}\psi(s_{i,j}, s_{i,j}(l))\lvert D(\mathbf{fd}_{i,j},l) - D(\mathbf{fs}_{i,j},l)\rvert, \\
		&D(\mathbf{fd}_{i,j},l)=\exp[-\frac{1}{2}(\mathbf{fd}_{i,j}\!-\!\mathbf{fd}_{i,j}(l))^{T}\mathbf{\Sigma}^{-1}_\mathbf{fd}(\mathbf{fd}_{i,j}\!-\!\mathbf{fd}_{i,j}(l))],
	\end{split}
\end{equation}
where $s_{ij}$ denotes the semantic label of ground truth. $\mathbf{fd}_{i,j}(l)$ is the depth feature which has an offset of $l$ along $x$ or $y$ direction and $\mathbf{fs}_{i,j}(l)$ is the semantic feature.  We use the exponential form of Mahalanobis Distance to measure the discrepancies between features where the covariance matrix $\mathbf{\Sigma}_{\mathbf{fd}}$ is set as a diagonal matrix $\sigma^{2}\mathbf{I}_{C}$. Here $\sigma$ is a learned parameter from each feature map. Considering that the depth features at the inner edges vary widely while the semantic representations are similar, we weight $L_{con}$ by the function $\psi(\bigcdot)$ which returns $1$ when the corresponding labels are different and $0$ otherwise. Since it is not realistic to take account of all the feature relationships, we select $l$ from the set $\{1,2\}$ so that each feature would be compared for $8$ times, which has adequately good performance in our experiments.

\textbf{Depth Loss:} The depth loss is composed of three items $\mathcal{L}_{berhu}$, $\mathcal{L}_{pair}$ and $\mathcal{L}_{norm}$.The $\mathcal{L}_{berhu}$ represents the BerHu Loss providing a good balance of the L1 norm and L2 norm, which is effective in the occasion errors following a heavy-tailed distribution \cite{Laina_2016}. The $\mathcal{L}_{berhu}$ is defined by
\begin{equation}
	\mathcal{L}_{berhu}=\sum_{i,j}
	\begin{cases}
		\lvert d_{i,j}-\tilde{d}_{i,j} \rvert & \text{if} \lvert d_{i,j}-\tilde{d}_{i,j}\rvert \leq c,\\
		\frac{(d_{i,j}-\tilde{d}_{i,j})^{2}+c^{2}}{2c} & \text{if} \lvert d_{i,j}-\tilde{d}_{i,j}\rvert > c
	\end{cases},
\end{equation}
where $d_{i,j}$ and $\tilde{d}_{i,j}$ are respectively the true and estimated depth values. $c$ is a threshold and we set it to be $c=\frac{1}{5}max_{k}(\lvert d_{k}-\tilde{d}_{k}\rvert)$, that is 0.2 times of the max error in a batch.

We also introduce the loss term $\mathcal{L}_{pair}$ to ensure the smoothness in the homogeneous regions and the relative distance of different areas. The formulation of $\mathcal{L}_{pair}$ is
\begin{equation}
	\mathcal{L}_{pair}=\sum_{p,q \in \Lambda, p \neq q }\lvert (d_{p}-d_{q}) - ( \tilde{d}_{p} - \tilde{d}_{q}) \rvert,
\end{equation}
in which $\Lambda=\{(i_{1},j_{1}),(i_{2},j_{2}),\dots,(i_{n},j_{n})\}$ denotes the set of  pixel indices which are selected randomly. We argue that $\mathcal{L}_{pair}$ not only keeps the advantages of the gradient loss \cite{Chen_2020} which penalizes the adjacent pixels of smooth areas and discontinued borders, but also provides similarity of pixels that are far apart, guaranteeing relative distances of different objects.

Another loss term is $\mathcal{L}_{norm}$, which is employed to emphasize small-size structures and high-frequency details:
\begin{equation}
	\mathcal{L}_{norm}=\sum_{i,j}1-\frac{\mathbf{n}_{i,j} \bigcdot \tilde{\mathbf{n}}_{i,j}}{\lvert \mathbf{n}_{i,j}\rvert \bigcdot \lvert \tilde{\mathbf{n}}_{i,j}\rvert},
\end{equation}
where $\mathbf{n}_{i,j}$ is the surface normal calculated by $\mathbf{n}_{i,j}=(-\nabla_{x}d_{i,j},-\nabla_{y}d_{i,j},1)^{T}$
, in which $\nabla_{x}$ and $\nabla_{y}$ represents the gradient values along the $x$-axis and $y$-axis separately. The depth loss is then calculated by the weighted summation of these three terms:
\begin{equation}
	\mathcal{L}_{depth}=\mathcal{L}_{berhu}+\lambda\mathcal{L}_{pair}+\mu\mathcal{L}_{norm},
\end{equation}
where $\lambda$,$\mu$ are weights to balance the depth loss terms.

\textbf{Segmentation Loss:} To ensure the accuracy of semantic segmentation and avoid unfavorable depth estimation along object boundaries, we introduce the weighted cross-entropy loss as segmentation loss $\mathcal{L}_{seg}$:
\begin{equation}
	\mathcal{L}_{seg}=-\sum_{i,j}\sum_{c}\omega_{c}\psi(s_{i,j},c)log(p(\tilde{s}_{i,j},c)),
\end{equation}
where $\omega_{c}=\frac{N_{total}-N_{c}}{N_{total}}$ weights the segmentation loss to mitigate the category imbalance problem. $p(\tilde{s}_{i,j},c)$ is the predicted probability value of class $c$. Then our total loss is
\begin{equation}
	\mathcal{L}=\mathcal{L}_{att}+\alpha \mathcal{L}_{depth}+\beta \mathcal{L}_{con}+\gamma \mathcal{L}_{seg},
\end{equation} 
where $\alpha$,$\beta$,$\gamma$ denotes the weight coefficients for each loss.

\section{Experiments}
In this section, we first introduce the training datasets and evaluation metrics.Then we illustrate the implementation details of training our model. Next, we investigate the effectiveness of our proposed network and compare it with other methods. Ablation study is also performed to show the benefits of our proposed methods.

\subsection{Dataset and Metrics}
\textbf{Dataset:} We use NYU-Depth-v2 \cite{Silberman_2012} and SUN-RGBD \cite{Song_2015} datasets to evaluate our presented model. NYU-Depth-v2 includes around 120K RGB-D images of 464 indoor scenes where only 1449 images from them have semantic labels. We follow the methods adopted in \cite{Jiao_2018,He_2021} to use the standard 795 training pairs and 654 testing pairs. SUN-RGBD is another dataset obtained from the indoor scenes containing about 10K images(5285 images for training and 5050 images for testing). Since the dataset has both semantic labels and depth labels, the whole training set is utilized to train our model and the test set to evaluate our semantic predictions. Both of these two datasets are employed for comparing with the other methods.

\textbf{Metrics:} Like the previous works \cite{eigen2014depth}, we assess our predicted depth maps using the following metrics:

Accuracy with threshold ($\delta^{p}$):
$\%$ of $d_{i,j}$ s.t. $max(\frac{\tilde{d}_{i,j}}{d_{i,j}},\frac{d_{i,j}}{\tilde{d}_{i,j}})=\delta^{p} < 1.25^{p}   (p=1,2,3)$

RMSE (rms): $\sqrt{\frac{1}{N}\sum_{i,j}(d_{i,j}-\tilde{d}_{i,j})^{2}}$

RMSE in log space (rms\_log): $\sqrt{\frac{1}{N}\sum_{i,j}(\ln d_{i,j}-\ln \tilde{d}_{i,j})^{2}}$

Mean absolute relative error (abs\_rel): $\frac{1}{N}\sum_{i,j}\frac{\lvert d_{i,j}-\tilde{d}_{i,j} \rvert}{d_{i,j}}$

Mean relative square error (sq\_rel): $\frac{1}{N}\sum_{i,j}\frac{(d_{i,j}-\tilde{d}_{i,j} )^{2}}{d_{i,j}^{2}}$

The signal $N$ represents the number of valid pixels.

To evaluate the predictions of semantic segmentation, we refer to the recent works \cite{Yu_2020,Deng_2015} and introduce pixel accuracy (pAcc) and mean intersection over union (mIoU) as metrics.

\subsection{Implementation Details}
We implement the model using the open source machine learning framework Pytorch on a single Nvidia GTX1080Ti GPU. As for the encoder of CI-Net, we choose the ResNet-50 and ResNet-101 as the candidates and both of them are pretrained on the ImageNet classification task \cite{saxena2007learning}. The learning rates of the pretrained layers are set to be $10$ times smaller than the other layers. To avoid the overfitting problem, we adopt the data augmentation strategies, including random rotation, random scaling, random crop, random horizontal flip and random color jitter. The Optimization Algorithm we used is stochastic gradient descent (SGD) \cite{Bottou_2010} where we set the momentum as 0.9 and the weight decay as $5e^{-4}$. To guarantee computational efficiency and fully optimizing the network, we choose the number of set $\Lambda$ as 500 to compute the pair loss $\mathcal{L}_{pair}$. The weight coefficients $(\alpha,\beta,\gamma,\lambda,\mu)$ are set to $(1,5,0.3,1,5)$ respectively. The training process is divided into three stages. At first, we replace the SUM module with the ground truth attention maps and the model is trained using only $\mathcal{L}_{depth}$ and $\mathcal{L}_{seg}$ (epochs and learning rates are $(300,6e^{-4})$ for NYU-Depth-v2 and $(60,6e^{-4})$ for SUN-RGBD). During the second stage, the SUM is added to the model and $L_{att}$ participates in the training process (epochs and learning rates are $(200,2e^{-4})$ for NYU-Depth-v2 and $(40,3e^{-4})$ for SUN-RGBD). In the last stage, we employ all the loss costs to train the entire model (we use the polynomical decay strategy with decayed power of 0.9, epochs and initial learning rates are $(200,2e^{-4})$ for NYU-Depth-v2 and $(40,3e^{-4})$ for SUN-RGBD).

%Notably, since the depth and segmentation features are confused at the beginning ot training, using consistency loss to optimize them is not effective. So we define the weight $\beta$ as a dynamic hyperparameter with $\beta=\max(\frac{2E_{cur}}{E_{toral}},1)$, where $E_{cur}$ and $E_{total}$ denote the current and the total epoch number. During the training phase, we train the model end-to-end with 300 epochs using the selected 12K images from NYU-Depth-v2 dataset and fine-tune it of 795 finely annotated images with 200 epochs. For the SUN-RGBD dataset, we train the model for 100 epochs. 

\subsection{Ablation Study}
In this subsection, we conduct numerous ablative experiments to analyze the effectiveness of our settings for the model. The experiments are evaluated on NYU-Depth-v2 dataset. To make an ablation study, we set a baseline network comprised of one encoder followed by two separate decoders each of which corresponds to a single task and consists of three blocks. On the basis of baseline, we also train a network with FSM to evaluate the effectiveness of SUM and FSM. Both networks and our proposed model are equipped with dilated ResNet-101 as encoder. 

\renewcommand\arraystretch{1.3}
\begin{sidewaystable}[htbp]
	\sidewaystablefn%
	\centering
	\caption{Ablation study of our method on the NYU-Depth-v2 dataset. The baseline model is our model removed FSM and SUM. Both depth estimation and semantic segmentation results are presented.}
	\begin{tabular}{c|ccc|c|ccc|cc|c|cc}
		\toprule
		\multirow{3}{*}{\textbf{Backbone}} & \multicolumn{3}{c|}{\textbf{Improvements}}&& \multicolumn{5}{c|}{\textbf{Depth}} && \multicolumn{2}{c}{\textbf{Segmentation}}     \\
		\cline{2-13} 
		& \multirow{2}{*}{SUM} & \multirow{2}{*}{FSM} & \multirow{2}{*}{$\mathcal{L}_{csc}$} && \multicolumn{3}{c|}{higher is better} & \multicolumn{2}{c|}{lower is better} &  & \multicolumn{2}{c}{higher is better} \\ 
		&&&&& $\delta_{1}$ & $\delta_{2}$ & $\delta_{3}$ & rms / rms\_log & abs\_rel / sq\_rel && pAcc & mIoU \\
		\hline
		\hline
		\multirow{5}{*}{Dilated ResNet-101} &&&&& 0.769 & 0.942 & 0.982 & 0.563 / 0.208 & 0.159 / 0.121 && 67.1 & 37.2 \\
		& $\checkmark$ &&&& 0.792 & 0.953 & 0.987 & 0.518 / 0.192 & 0.142 / 0.115 && 70.9 & 41.2 \\
		&& $\checkmark$ &&& 0.785 & 0.950 & 0.988 & 0.538 / 0.201 & 0.152 / 0.118 && 69.5 & 39.0 \\
		& $\checkmark$ & $\checkmark$ &&& 0.809 & 0.957 & 0.990 & 0.512 / 0.187 & 0.140 / 0.112 && 72.2 & 42.5 \\
		& $\checkmark$ & $\checkmark$ & $\checkmark$ && 0.812 & 0.957 & 0.990 & 0.504 / 0.181 & 0.129 / 0.112 && 72.7 & 42.6 \\
		\bottomrule
	\end{tabular}
	\label{qualitative results of ablation study}
\end{sidewaystable}

\begin{figure*}[htbp]
	\centering
	\setlength{\abovecaptionskip}{0.5cm}
	\begin{overpic}[width=4.7in]{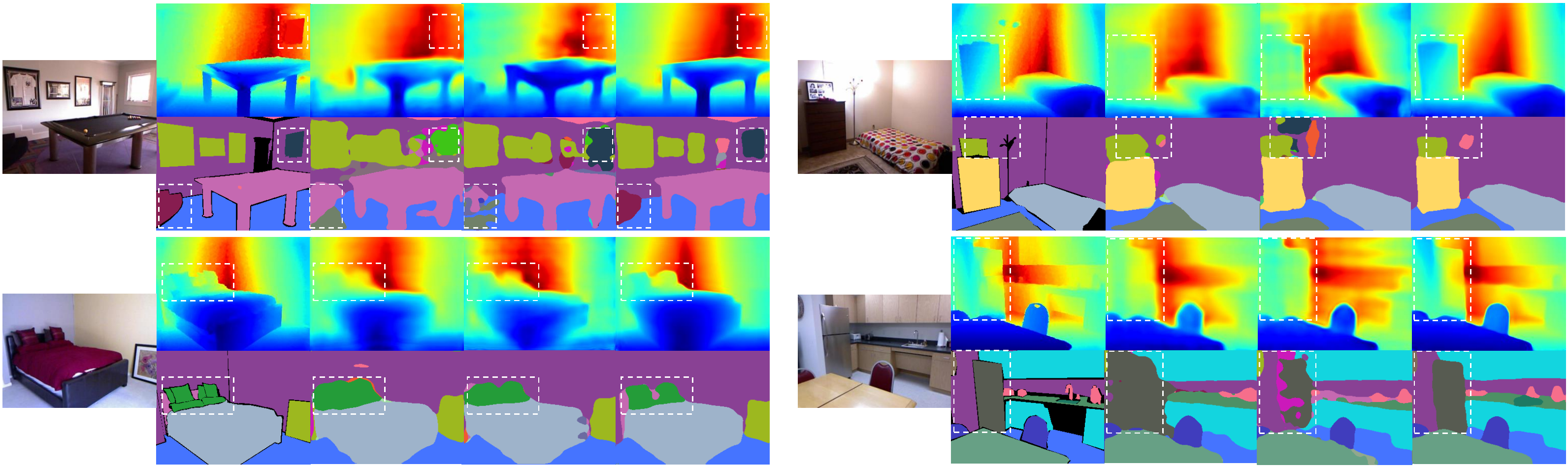}
		\put(3,-4){(a)}
		\put(13,-4){(b)}
		\put(23,-4){(c)}
		\put(33,-4){(d)}
		\put(43,-4){(e)}
		\put(54,-4){(a)}
		\put(64,-4){(b)}
		\put(74,-4){(c)}
		\put(84,-4){(d)}
		\put(94,-4){(e)}
	\end{overpic}
	\caption{Ablative visual comparisons. (a) input image; (b) ground truth; (c) results of baseline; (d) results of baseline with FSM; (e) results of our method.}
	\label{visual results of ablation study}
\end{figure*}
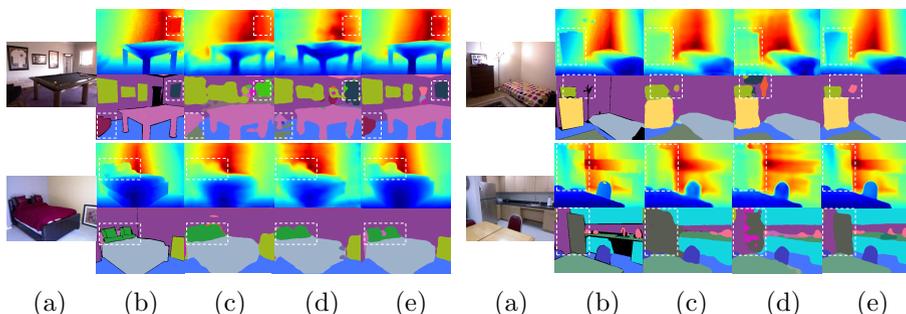

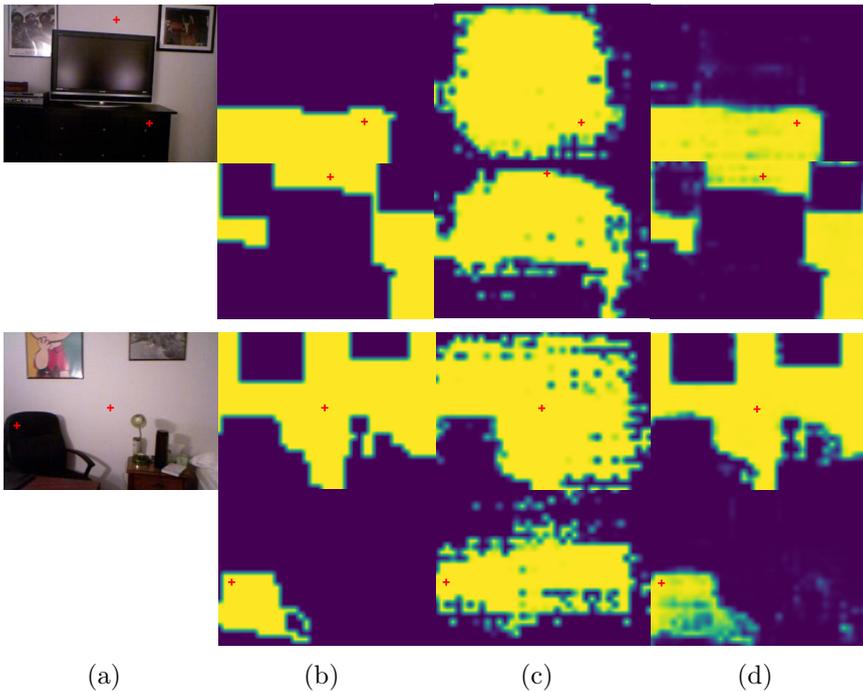
\begin{figure}[htbp]
	\centering
	\setlength{\abovecaptionskip}{0.5cm}
	\begin{overpic}[width=4.5in]{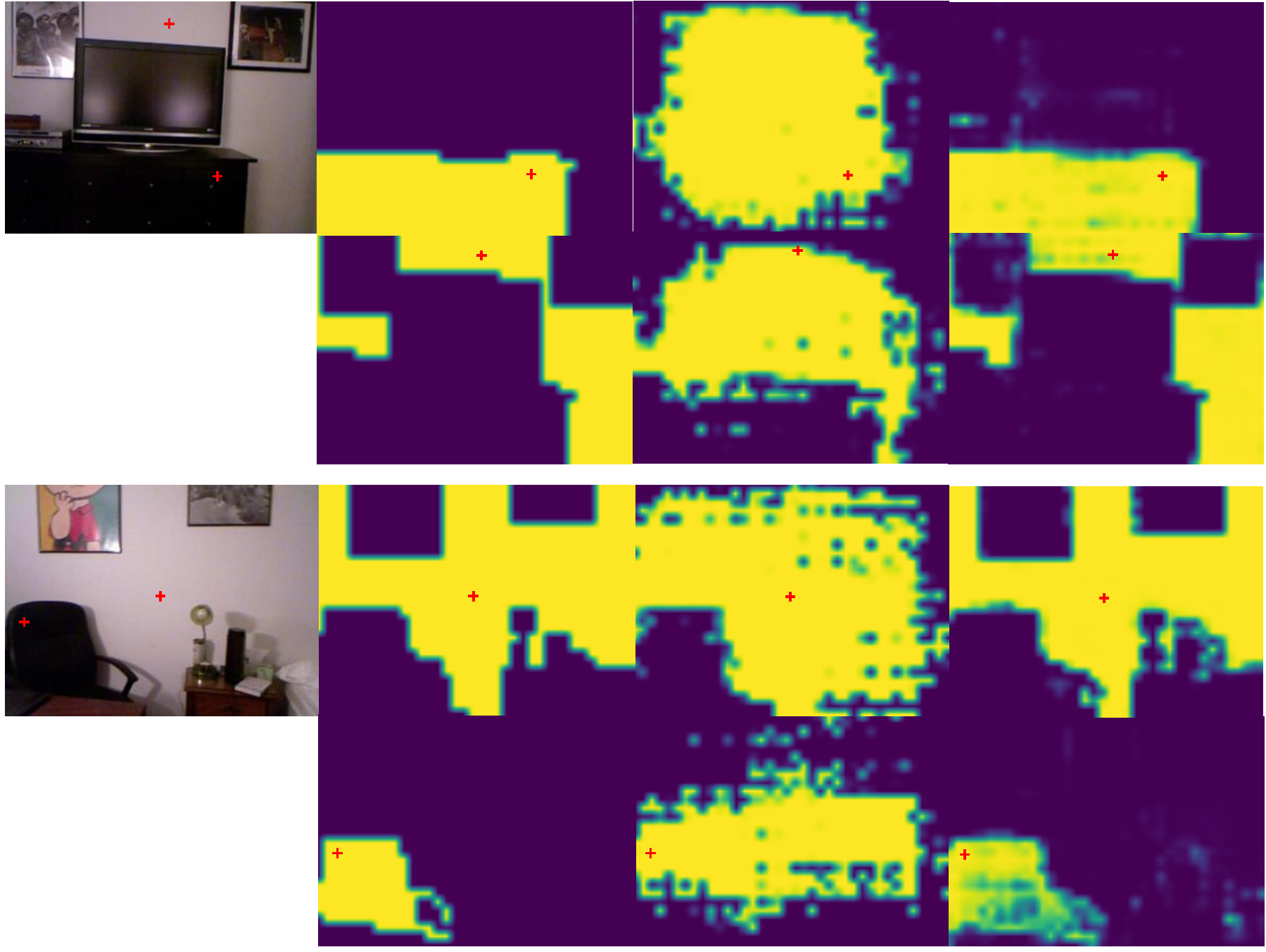}
		\put(10,-4){(a)}
		\put(35,-4){(b)}
		\put(60,-4){(c)}
		\put(85,-4){(d)}
	\end{overpic}
	\caption{Comparisons of the learned attention map. (a) input images; (b) ideal attention maps; (c) and (d) represent the attention maps produced by our model without and with supervision of the $\mathcal{L}_{att}$ respectively. For each scene we show two different attention maps pertain to the locations where the red plus signs mark.}
	\label{visual comparisons of attention maps}
\end{figure}

We first analyze the contribution of scene understanding module. The ablative visual results are shown in (d) and (e) of Figure \ref{visual results of ablation study}. It can be observed that without context prior, both the depth estimation and segmentation results of baseline with FSM suffer noticeable errors especially in the white dashed line boxes. We also observe that SUM can significantly reduce the adverse effect of uneven illumination. For example, in the second scene there is a lamp shedding intense light which impairs the baseline prediction of the surrounding region. In this case, SUM provides the understanding of scene, which helps accurately predict depth and semantic information. In addition, we visualize the attention maps with and without the supervision of attention loss respectively. Figure \ref{visual comparisons of attention maps} shows that with the guidance of attention loss, the model does capture the correlated contextual areas and adapts to different scenes well. The attention map with supervision could be regarded as a structural extractor since it extracts intact object shapes, revealing the layout of scene. Whereas for the attention maps are not properly guided, the resulting arbitrary concerned region can be harmful.

\renewcommand\arraystretch{1.3}
\begin{table}[htbp]
	\centering
	\caption{Comparisons with the depth estimation methods on NYU-Depth-V2 dataset.}
	\begin{tabular}{c|c|ccc|cccc}
		\toprule
		\multirow{2}{*}{\textbf{Methods}}  & \multirow{2}{*}{\textbf{Data}} & \multicolumn{3}{c|}{higher is better} & \multicolumn{4}{c}{lower is better} \\ \cline{3-9} 
		&& $\delta_{1}$ & $\delta_{2}$ & $\delta_{3}$ & rms & rms\_log & abs\_rel & sq\_rel \\
		\hline
		\hline
		Wang et al.\cite{Peng_Wang_2015}  & 795  & 0.605 & 0.890 & 0.970 & 0.745 & 0.262 & 0.220 & 0.210\\
		DCNF\cite{Liu_2015} & 795  & 0.614 & 0.883 & 0.971 & 0.824 & / & 0.230 & / \\
		HCRF\cite{Bo_Li_2015}  & 795  &  0.621 & 0.886 & 0.968 & 0.821 & / & 0.232 & / \\ 
		Roy et al.\cite{Roy_2016} & 795  & / & /& /  & 0.744 & / & 0.187 & / \\
		PAD-Net \cite{Xu_2018_Pad}  & 795 & 0.817 & 0.954 & 0.987 & 0.582 & / & 0.120 & / \\
		\hline
		FCRN \cite{Laina_2016}  & 12k & 0.811 & 0.953 & 0.988 & 0.573 & 0.195 & 0.127 & / \\
		Li et al. \cite{Li_2018} & 12k & 0.820 & 0.960 & 0.989 & 0.545 & / & 0.139 & / \\
		GeoNet \cite{Qi_2018}  & 16k & 0.834 & 0.960 & 0.990 & 0.569 & / & 0.128 & / \\
		\hline
		Eigen et al.\cite{eigen2014depth} & 120k & 0.611 & 0.887 & 0.971 & 0.907 & 0.285 & 0.158 & 0.121 \\
		Eigen et al.\cite{Eigen_2015} & 120k & 0.769 & 0.950 & 0.988 & 0.641 & 0.214 & 0.158 & 0.121 \\
		CRFs\cite{Xu_2018} & 95k  & 0.811 & 0.954 & 0.987 & 0.586 & / & 0.121 & / \\
		DORN \cite{Fu_2018} & 120k & 0.828 & 0.965 & 0.992 & 0.509 & / & 0.115 & / \\
		Hu et al. \cite{Hu_2019}  & 120k  & 0.866 & 0.975 & 0.993 & 0.530 & / & 0.115 & / \\
		\hline
		\hline
		CI-Net & 795 & 0.812 & 0.957 & 0.990 & 0.504 & 0.181 & 0.129 & 0.112 \\
		\bottomrule
	\end{tabular}
	\label{comparison of depth estimation results on NYU}
\end{table}

Next, it is obvious that the feature sharing strategy improves the performance of both semantic segmentation and depth estimation. For the example in the forth scene of Figure \ref{visual results of ablation study}, the baseline fails to predict the wall behind fridge while FSM helps to perceive the existence of these two objects and make boundaries in the depth map clearer. These improvements are facilitated by deeply fusing the task specific features, providing more robust information for prediction. 

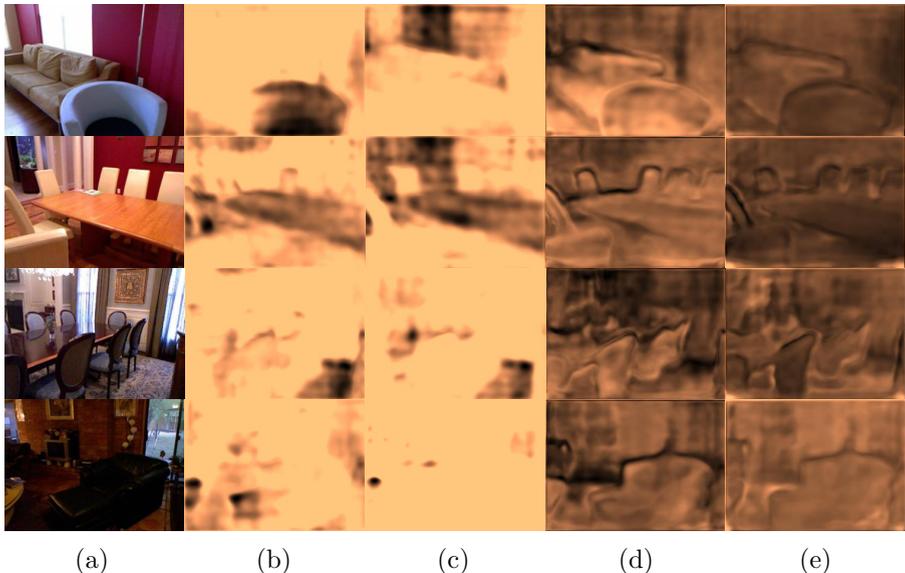
\begin{figure}[htbp]
	\centering
	\setlength{\abovecaptionskip}{0.5cm}
	\begin{overpic}[width=4.7in]{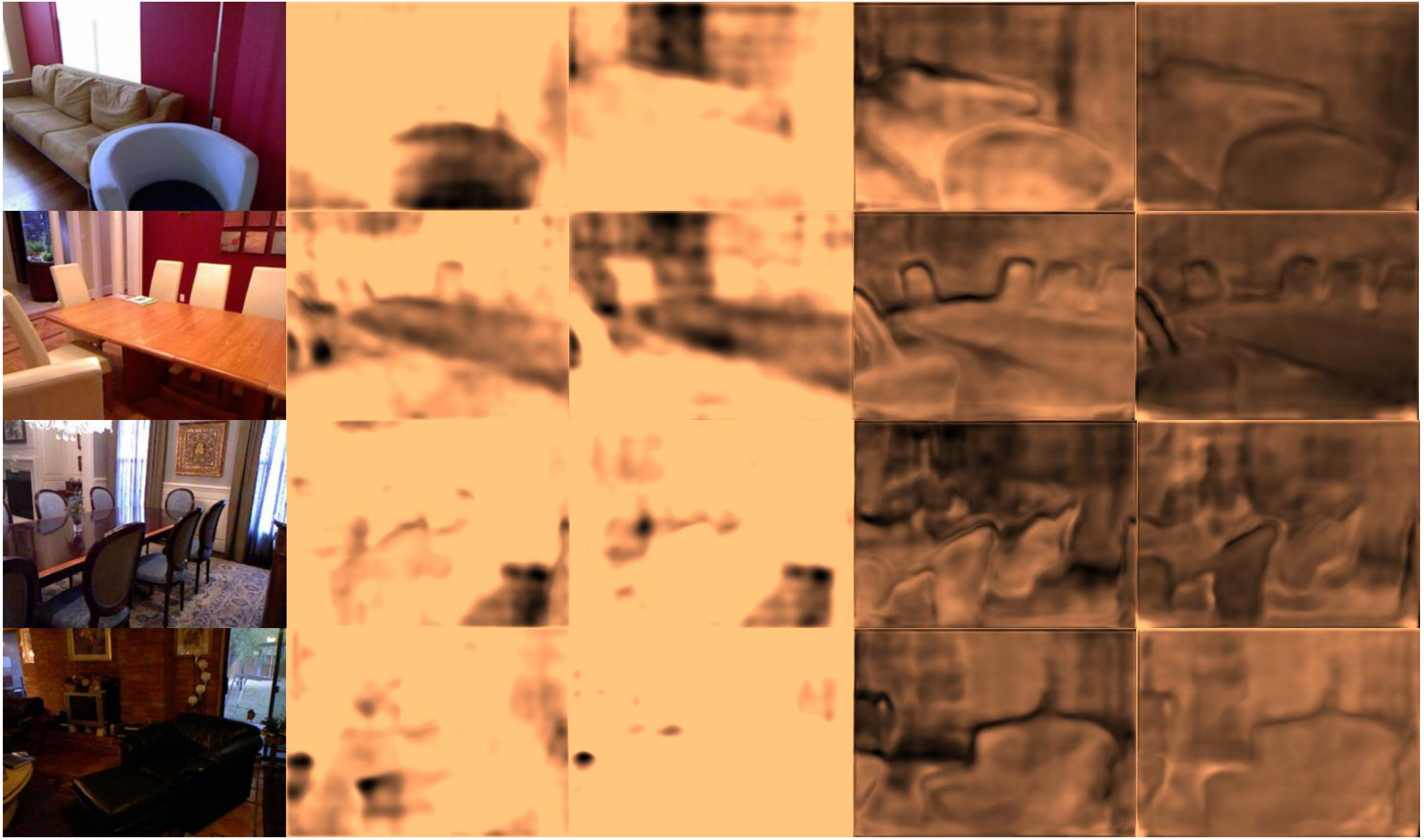}
		\put(8,-4){(a)}
		\put(28,-4){(b)}
		\put(48,-4){(c)}
		\put(68,-4){(d)}
		\put(88,-4){(e)}
	\end{overpic}
	\caption{Visual results of allocated features(summed along the channel dimension and normalized to $0-1$ range). (a) input images; (b) features added to semantic branch learned by LSU; (c) features added to depth branch learned by LSU; (d) features added to semantic branch learned by FSM (e) features added to depth branch learned by FSM.}
	\label{visual comparisons of LSU and FSM}	
\end{figure}

To see the improvement of SUM and FSM, we also perform quantitative comparisons, which can be seen in Table \ref{qualitative results of ablation study}. It can be obviously found that both the introductions of SUM and FSM improve the performance significantly, which verifies the effectiveness of these two blocks. For the case with scene understanding module, i.e., baseline with SUM, the original baseline network could obtain a prominent gain on both tasks, especially in rms and mIoU ($7.9\%$ reduced and $10.7\%$ increased respectively). This result agrees well with the qualitative data and indicates that improving the understanding of scene is effective. Moreover, we also exhibits that when the extra supervision of consistency loss is added, the accuracies are slightly enhanced.

\subsection{Comparisons with other methods}
In this subsection, we first analyze the differences between FSM and LSU quantitatively and qualitatively. Then we compare the experimental results of our model with other algorithms according to different tasks.

\textbf{Comparisons of LSU and FSM:} We test the effectiveness of FSM and LSU \cite{choi2020safenet,Jiao_2018} whose results can be seen in Table \ref{comparative study of FSM and LSU}. It is noticed that although the LSU does improve the performance, the improvements are not as obvious as those by adding FSM, which proves the effectiveness of sampling operations. To make a deep analysis of the difference, we visualize the allocated features $\mathcal{F}_{LSUd}$, $\mathcal{F}_{LSUs}$, $\mathcal{F}_{FSMd}$, $\mathcal{F}_{FSMs}$(summed long the channel dimension and normalized to $0-1$ range) of LSU and FSM, which are respectively formulated as:
\begin{equation}
	\begin{split}
		\mathcal{F}_{LSUd}&=C_{\mathbf{fd}1}^{1\times1}(\mathbf{fd}_{t}) +  C_{\mathbf{fs}2}^{1\times1}(\mathbf{fs}_{t}) \\
		\mathcal{F}_{LSUs}&=C_{\mathbf{fd}2}^{1\times1}(\mathbf{fd}_{t}) +  C_{\mathbf{fs}1}^{1\times1}(\mathbf{fs}_{t}) \\
		\mathcal{F}_{FSMd}&=C_{\mathbf{fd}}^{1\times1}(C(concat(\mathbf{fd}_{t},\mathbf{fs}_{t}))) \\
		\mathcal{F}_{FSMs}&=C_{\mathbf{fs}}^{1\times1}(C(concat(\mathbf{fd}_{t},\mathbf{fs}_{t})))
	\end{split}
\end{equation}
In Figure \ref{visual comparisons of LSU and FSM}, we can easily observe that the features learned by LSU almost pay attention to the whole region, which is not reasonable in depth estimation and semantic segmentation since if all the features of different objects are emphasized, the classification and depth estimation of objects would be confused for using highlighted features from others. In contrast, with larger receptive fields and deeper structure, our proposed module could learn more useful information such as objects, which are important in both tasks. In addition, it could be observed that FSM learns clear black boundaries between different objects, when the features are used for lateral convolution, such contours of zero values would inhibit the convoluted operation from using features of other objects, avoiding generating confused features.

\begin{table}[htbp]
	\centering
	\setlength\tabcolsep{3.5pt}
	\caption{Comparisons between FSM and LSU}
	\begin{tabular}{c|ccc|cc|cc}
		\toprule
		& $\delta_{1}$&$\delta_{2}$&$\delta_{3}$& rms & abs\_rel & pAcc & mIoU \\
		\hline
		\hline
		baseline & 0.769&0.942&0.982 & 0.563 & 0.159 & 67.1 & 37.2\\
		baseline + LSU & 0.778&0.948&0.986 & 0.547 & 0.152 & 68.3 & 38.5 \\
		baseline + FSM & 0.785&0.950&0.988 & 0.538 & 0.152 & 69.5 & 39.0\\
		\bottomrule
	\end{tabular}
	\label{comparative study of FSM and LSU}
\end{table}

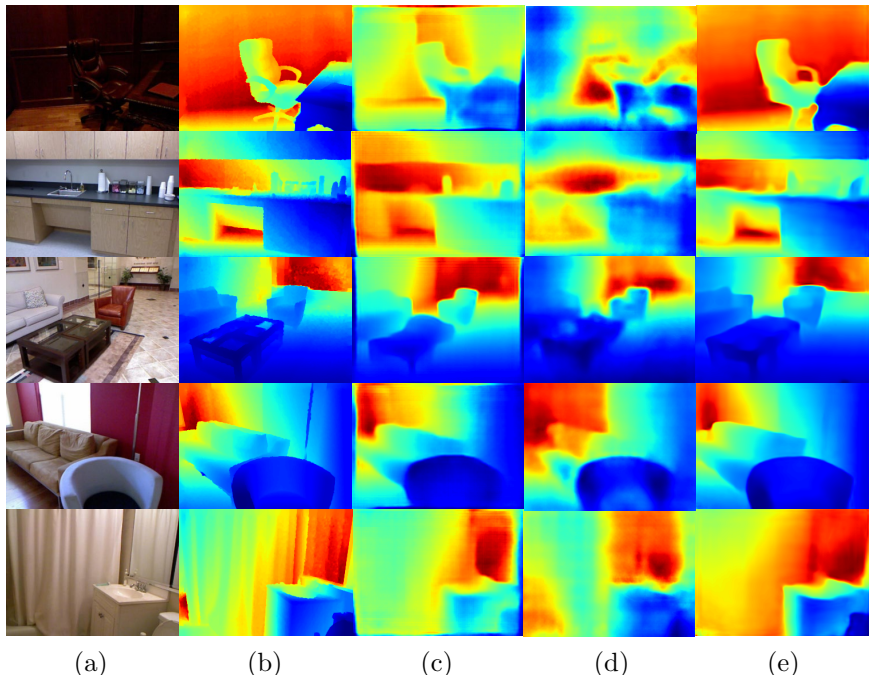
\begin{figure}[htbp]
	\centering
	\setlength{\abovecaptionskip}{0.5cm}
	\begin{overpic}[width=4.5in]{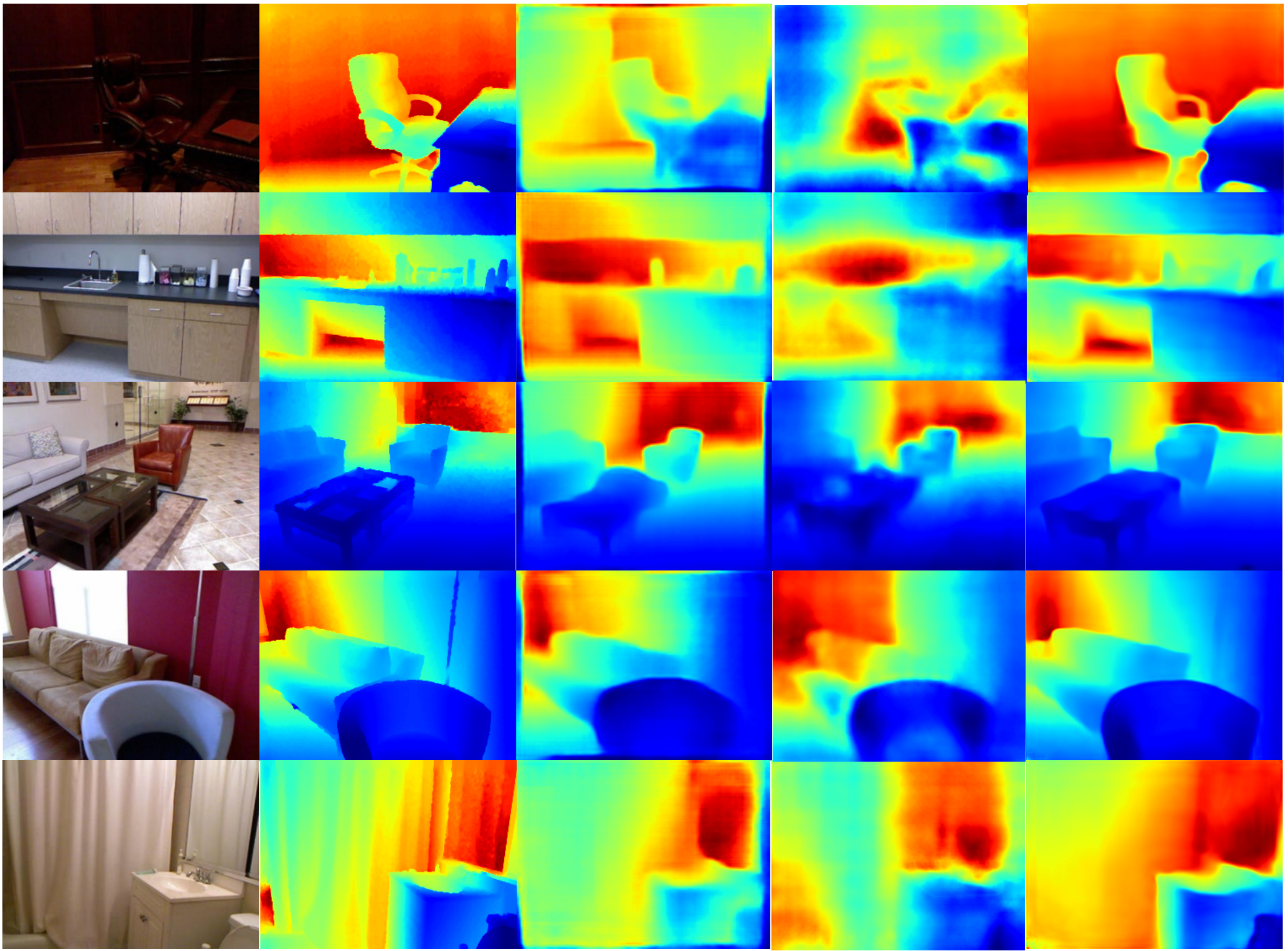}
		\put(8,-4){(a)}
		\put(28,-4){(b)}
		\put(48,-4){(c)}
		\put(68,-4){(d)}
		\put(88,-4){(e)}
	\end{overpic}
	\caption{Visual comparison with some approaches on NYU-depth-v2 dataset. (a) input images; (b) ground truth; (c) predictions of \cite{Laina_2016}; (d) predictions of \cite{Xu_2018} (e) predictions of our model.}
	\label{visual comparisons of depth estimation}
\end{figure}

\textbf{Depth Estimation:} Comparisons of depth values are made among our presented CI-Net and other algorithms on NYU-Depth-v2 dataset where the number of categories are 40. The results are summarized in Table \ref{comparison of depth estimation results on NYU}. The values are used directly from the original papers. As it can be seen in the Table \ref{comparison of depth estimation results on NYU}, our model with ResNet-101 achieves best results in most metrics compared to the model using the same scale data. The rms of CI-Net is even better than those methods fed with larger training set. It's also worth noting that although the methods in \cite{Fu_2018,Hu_2019,Xu_2018} perform well in part of the evaluating indicators, they all feed numerous images (120k) to boost the performance.

%and the ResNet-101 based model obtains superior outcomes in $\delta^{1}$, RMSE, sq\_rel and abs\_rel thanks to the designed SUM, FSM and consistency loss function. It is worth noting that although our proposed model is trained on 12K images, it even performs better than those methods using larger data. As for the segmentation results, contrast to the method in \cite{bibid}, our ResNet-50 based model has a slightly gain for partial metrics while the network with stronger backbone improves considerably for pixel-acc and mIoU.

\begin{table}[htbp]
	\centering
	\caption{Comparisons with the semantic segmentation methods on NYU-Depth-v2 dataset}
	\begin{tabular}{c|c|cc}
		\toprule
		\textbf{Methods} & \textbf{Data} & pAcc & mIoU \\
		\hline
		\hline
		FCN \cite{Long_2015} & RGB & 60.0 & 29.2\\
		Context \cite{Lin_2016} & RGB & 70.0 & 40.6 \\
		Eigen et al. \cite{Eigen_2015}& RGB & 65.6 & 34.1 \\
		B-SegNet \cite{kendall2015bayesian} & RGB & 68.0 & 32.4 \\
		RefineNet-LW-101 \cite{nekrasov2018light}& RGB & / & 43.6 \\
		TD2-PSP50 \cite{Hu_2020} & RGB & / & 43.5 \\
		Deng et al. \cite{Deng_2015}& RGBD & 63.8 & 31.5 \\
		3DGNN \cite{Qi_2017} & RGBD & / & 42.0 \\
		D-CNN \cite{Wang_2018} & RGBD & / & 41.0 \\
		\hline
		\hline
		CI-Net & RGB & 72.7 & 42.6 \\ 
		\bottomrule
	\end{tabular}
	\label{comparison results of semantic segmenation on NYU-Depth-v2 dataset}
\end{table}

In addition, Figure \ref{visual comparisons of depth estimation} displays some visual examples. It can be seen that though the predictions of the method by Laina \cite{Laina_2016} are smoothed as a whole, they lose some details, bringing in the blurred object boundaries especially in desk, washing machine and sofa. Besides, the precision of depth maps is weak, the depths of some regions deviate the ground truth severely. Although \cite{Xu_2018} has great values in evaluation metrics as seen in Table \ref{comparison of depth estimation results on NYU}, the contours in predicted depth maps of their models are not sharp which make the depth maps not clear enough. Compared to them, our results match the structures of scenes and have sharper object boundaries benefiting from prior inter-class and intra-class information. 

\begin{table}[htbp]
	\centering
	\caption{Comparisons with the semantic segmentation methods on SUN-RGBD dataset}
	\begin{tabular}{c|c|cc}
		\toprule
		\textbf{Methods} & \textbf{Data} & pAcc & mIoU \\
		\hline
		\hline
		Context \cite{Lin_2016} & RGB & 78.4 & 42.3 \\
		B-SegNet \cite{kendall2015bayesian} & RGB & 71.2 & 30.7 \\
		RefineNet-101 \cite{Lin_2017} & RGB & 80.4 & 45.7 \\
		AdaptNet++ \cite{Valada_2019} & RGB & / & 38.4 \\
		SSMA \cite{Valada_2019} & RGBD & 80.2 & 43.9 \\
		CMoDE \cite{Valada_2017_Deep} & RGBD & 79.8 & 41.9 \\
		LFC \cite{Valada_2017} & RGBD & 79.4 & 41.8 \\
		FuseNet \cite{Hazirbas_2017} & RGBD & 76.3 & 37.3 \\
		D-CNN \cite{Wang_2018_Depth} & RGBD & / & 42.0 \\
		\hline
		\hline
		CI-Net & RGB & 80.7 & 44.3 \\ 
		\bottomrule
	\end{tabular}
	\label{comparison results of semantic segmenation on SUN-RGBD dataset}
\end{table}

\begin{figure}[htbp]
	\centering
	\setlength{\abovecaptionskip}{0.5cm}
	\begin{overpic}[width=4.0in]{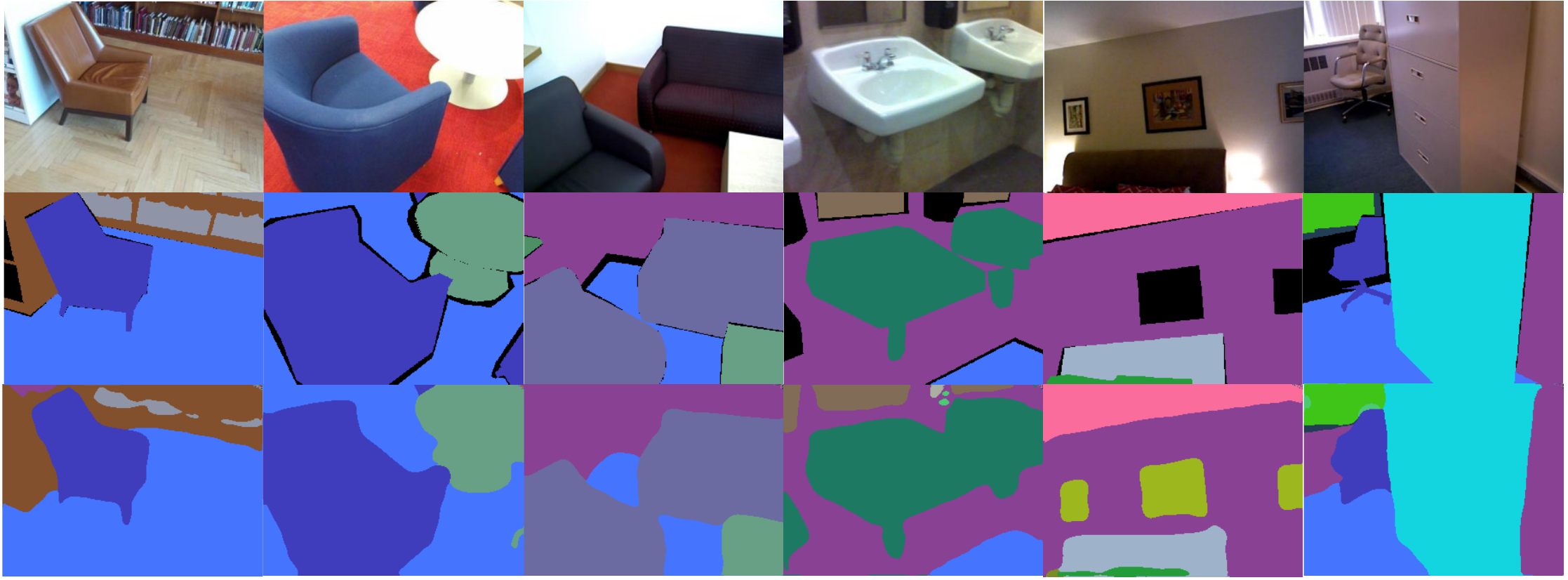}
		\put(-6,29){(a)}
		\put(-6,17){(b)}
		\put(-6,6){(c)}
	\end{overpic}
	\caption{Visualized semantic segmentation maps on SUN-RGBD dataset. (a) input images; (b) ground truth; (c) semantic segmentation predictions of our method.}
	\label{visual semantic results of SUN-RGBD}
\end{figure}

\textbf{Semantic Segmentation:} Apart from depth estimation, we also show the semantic segmentation results in Table \ref{comparison results of semantic segmenation on NYU-Depth-v2 dataset} and Table \ref{comparison results of semantic segmenation on SUN-RGBD dataset} for NYU-Depth-v2 and SUN-RGBD datasets respectively. The data types of RGB and RGBD in tables mean using RGB images or both RGB and depth images as input. As reported in Table \ref{comparison results of semantic segmenation on NYU-Depth-v2 dataset}, our model with ResNet-101 performs best among all the listed methods in both pixel-acc and mIoU even though our method only uses RGB images, which demonstrate that the context prior can also benefit the semantic segmentation and the deep feature interaction does help this task leverage more useful information. As for the results of SUN-RGBD dataset in Table \ref{comparison results of semantic segmenation on SUN-RGBD dataset}, we can notice that our method still performs competitively. Although RefineNet-101 achieves best performance in mIoU metric, our method is on a par with it in pixel-acc metric. Some selected  visual maps of both datasets are also shown in Figure \ref{visual semantic results of SUN-RGBD}, it can be seen that our predictions are in high quality and even give the correct labels of the invalid regions in ground truth.

\section{Conclusion}\label{sec13}

In this paper, a network for joint task learning was proposed. By employing the scene understanding module, the presented network was able to capture the contextual information of inter-class and intra-class, which is crucial for the network to understand which useful context can be exploited to make predictions. To fuse the task-specific features deeply, we designed a feature sharing module which enlarged the receptive fields and augmented the presentation ability through upsampling and downsampling operations. In addition, a consistency loss was proposed to make the task-specific features mutually guided, which kept consistent relationships with the respective adjacent features. Ablation study demonstrated the improvements of our proposed modules and the comparative results with other methods on NYU-Depth-v2 and SUN RGB-D datasets the effectiveness of our method. In the future, we plan to explore the contextual information in the attention map and incorporate other pixel-level tasks such as surface normal prediction, edge detection into this work. Also, we are interested in combining the depth prediction with semantic SLAM to obtain more accurate results.

\section*{Declarations}

\begin{itemize}
	\item Funding: Not Available
	\item Conflict of interest/Competing interests (check journal-specific guidelines for which heading to use): Not Available
	\item Availability of data and materials: Not Available
	\item Code availability: Not Available
\end{itemize}


\bibliography{references}% common bib file


\end{document}